\documentclass{article}

\usepackage{arxiv}

\usepackage[utf8]{inputenc} 
\usepackage[T1]{fontenc}    
\usepackage{hyperref}       
\usepackage{url}            
\usepackage{booktabs}       
\usepackage{amsfonts}       
\usepackage{nicefrac}       
\usepackage{microtype}      
\usepackage{lipsum}
\usepackage{multirow}
\usepackage{makecell}

\usepackage{times} 
\usepackage{helvet}
\usepackage{courier} 
\usepackage{url} 
\usepackage{graphicx} 
\usepackage{epsfig}
\usepackage{amsmath}
\usepackage{amssymb}
\usepackage{pdflscape}
\usepackage{tabularx}
\usepackage{lipsum}
\usepackage{multicol}
\usepackage{multirow}
\usepackage{array}
\usepackage{stfloats}

\newcommand\blfootnote[1]{%
  \begingroup
  \renewcommand\thefootnote{}\footnote{#1}%
  \addtocounter{footnote}{-1}%
  \endgroup
}

\title{Rethinking Irregular Scene Text Recognition}

\author{
  Shangbang Long$^\dagger$ \\
  Department of Machine Learning \\
  Carnegie Mellon University\\
  Pittsburgh, USA \\
  \texttt{longshangbang@cmu.edu} \\
   \And
 Yushuo Guan \\
  School of EECS\\
  Peking University\\
  Beijing, China \\
  \texttt{david.guan@pku.edu.cn} \\
   \And
 Bingxuan Wang \\
  Yuanpei College\\
  Peking University\\
  Beijing, China \\
  \texttt{wangbx@pku.edu.cn} \\
   \And
 Kaigui Bian \\
  School of EECS\\
    Peking University\\
  Beijing, China \\
  \texttt{bkg@pku.edu.cn} \\
  \And
 Cong Yao \\
   MEGVII (Face++) Inc.\\
  Beijing, China \\
  \texttt{yaocong2010@gmail.com} \\
}

\begin{document}
\maketitle

\begin{abstract}
Reading\blfootnote{$^\dagger$Work done while Shangbang Long was an intern at MEGVII (Face++) Inc., Beijing, China. } text from natural images is challenging due to the great variety in text font, color, size, complex background and etc.. 
The perspective distortion and non-linear spatial arrangement of characters make it further difficult. 
While rectification based method is intuitively grounded and has pushed the envelope by far, its potential is far from being well exploited. 
In this paper, we present a bag of techniques that prove to significantly improve the performance of rectification based method. 
On curved text dataset, our method achieves an accuracy of $89.6\%$ on CUTE-80 and $76.3\%$ on Total-Text, an improvement over previous state-of-the-art by $6.3\%$ and $14.7\%$ respectively. 
Furthermore, our combination of techniques helps us win the ICDAR 2019 Arbitrary-Shaped Text Challenge (Latin script), achieving an accuracy of $74.3\%$ on the held-out test set. 
We release our code as well as data samples for further exploration at \url{https://github.com/Jyouhou/ICDAR2019-ArT-Recognition-Alchemy}. 
\end{abstract}


\section{Introduction}
Recently, the detection and recognition of irregular text from natural images have become a new popular research topic~\cite{long2018scene,Shi_2017_CVPR,Zhou_2017_CVPR,Deng2018,long2018textsnake,baek2019character,tian2019learning,shi2016robust,liao2018scene,li2018show}. 
However, most detection methods describe text as bounding boxes or groups of pixels in the form of semantic segmentation, without any indication of the shapes. 
Thus, the recognition models still need proper mechanism to deal with curved text. 

Deep learning based text recognition algorithms have transitioned from \textit{one-dimensional} methods, i.e. CRNN and CTC based methods~\cite{su2014accurate,liu2016star,lee2016recursive,cheng2017focusing,shi2017end,gao2017reading,yin2017scene}, to \textit{quasi-two-dimensional} methods using rectification~\cite{shi2016robust,shi2018aster} which is adapted fro Spatial Transformer Networks (STN)~\cite{jaderberg2015spatial}, and currently to \textit{two-dimensional} methods with Fully Convolutional Networks~\cite{long2015fully} or spatial attention~\cite{xu2015show} such as~\cite{liao2018scene,li2018show}. 
These methods have started and followed a standard and well accepted training and evaluation protocol, using images rendered with synthesis engines~\cite{jaderberg2014synthetic,gupta2016synthetic} as training data and then evaluating on real world datasets.  
Besides, they share a standard pre-processing step by resizing  all images to a fixes size, usually set as $64\times256$ pixels. 

Despite these widely-adopted trends, the aforementioned common practices fall short when faced with irregular text. 
Specifically, synthetic data generated by existing algorithms~\cite{jaderberg2014synthetic,gupta2016synthetic,zhan2018verisimilar,liao2019synthtext3d} largely consist of text aligned to straight lines. 
Although the 800K synthetic images generated by SynthText~\cite{gupta2016synthetic} do contain curved text, the proportion is small and they are only slightly curved. 
Besides, images with irregular text are usually seriously warped if they are resized to $64\times256$. 
Therefore, it is important and worthwhile to study the bottleneck effect of these factors in building strong and robust text recognizers. 

In this paper, we examine a set of techniques from different aspects that improve the performance of recognizers with focus on irregular text. 
These techniques may seem to be slight modification to models, data, and training procedures at first glance. 
However, they are reasonable and bring considerable improvement over baselines and even beat state-of-the-art models by a large margin. 
We design comprehensive experiments to analyse the effect of these techniques. 
Furthermore, we are also the first to evaluate on the newly published curved text dataset, Total-Text~\cite{kheng2017total}, which contains $2201$ images, much larger than the CUTE80 dataset~\cite{risnumawan2014robust}. 
Images in Total-Text are characterized by even more challenging conditions, and thus should be a better performance indicator for curved text detectors. 

Our paper is organized as follows. 
In Section \ref{baseline}, we first set up our baseline methods and experiment settings. 
The experiments throughout this paper follow the same settings for fair comparison, unless otherwise specified. 
In Section \ref{data}, we discuss the selection of training data. 
In Section \ref{tweak}, we evaluate several model modifications. 
In Section \ref{ic19}, we review our participation in the ICDAR 2019 ArT challenge, and describe our system used in the competition. 
In Section \ref{conc}, we summarize our findings and contributions. 

The code and training scripts are all included in our Github repository as shown in the abstract.


\section{Baseline System and Experiment Settings} \label{baseline}
Basically, we follow the standard of training and evaluating scene text recognition algorithms as most previous works do. 
We give a brief description of our baseline recognition system in Section \ref{bl:model}. We choose a rectification based method as our experiment tools. 
Then, we specify our experiments settings in Section \ref{bl:train} and Section \ref{bl:eva}. 
Details regarding to the models and experiments can be found in our code repository. 
Especially, we would like to thank Mingkun Yang for his repository\footnote{\url{https://github.com/ayumiymk/aster.pytorch}} of the Aster algorithm~\cite{shi2018aster}.

\subsection{Baseline Models with Rectification Layer} \label{bl:model}

\begin{figure}
\centering
\includegraphics[width=0.8\textwidth]{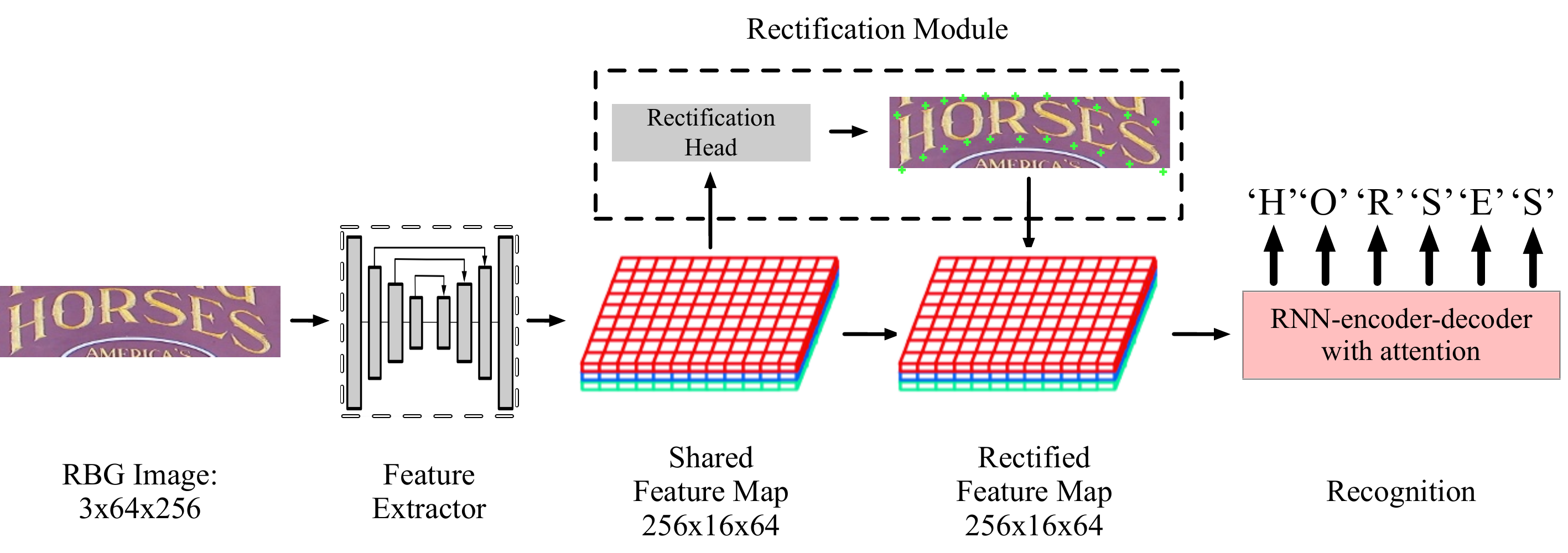}

\caption{The pipeline of our implementation. The module marked with dotted rectangles is the rectification module. When we take off the rectification module, it becomes a vanilla RNN and CNN based model. The green dots inside represent predicted control points. For implementation details, we refer readers to our code. }  
\label{fig:pipeline}
\vspace{-4mm}
\end{figure}

We follow and modify the PyTorch implementation of Aster~\cite{shi2018aster} released by its authors as mentioned above. 
The pipeline is shown in Fig. \ref{fig:pipeline}, and is composed of the following steps:

\noindent \textbf{Preprocessing} Input images are first resized to $64\times256$. 
Pixel values are subtracted and then divided by $128.0$, thus normalized to a $-1.0$ to $0.99$ base. 

\noindent \textbf{Feature Extractor} We use a standard 50-layered ResNet~\cite{He_2017_Res} as backbone network and Feature Pyramid Networks~\cite{Lin_2017_CVPR} to extract features from images. 
The output feature maps have $\frac{1}{4}$ the resolution of input images, that is, $16\times64$, and $256$ channels. 
We believe that the widely-used ResNet+FPN feature extractor should serve as a fair playground for easy replication and controlled experiments. 

\noindent \textbf{Rectification Module} To perform rectification, we first apply a control point header to localize the text with a $20$-vertexed polygon. 
The header consists of a series of alternating convolutional layers and maxpooling layers that encode the $256\times16\times64$ feature maps from feature extractor to a $512$-d vector. 
Two fully connected networks follow and output the polygon. 
Then, the feature maps are rectified to axis-aligned form with Thin-Plate-Spline (TPS)~\cite{warps1989thin}. 
Details with regard to the grid generator is the same as that specified in Aster. 

\noindent \textbf{Recognition Module} The input to the recognition modules is the rectified feature maps output from the rectification module, with a size of $256\times16\times64$. 
We follow the PyTorch implementation of Aster, and attach several convolutional layers, maxpooling layers and Bi-LSTM layers~\cite{hochreiter1997long} layers which act as encoders, and then attentional decoders. 
The decoder is an LSTM with attention mechanism to predict a sequence of symbols from a predefined character list. 
To produce sequence of variable length, a special end-of-sentence symbol, denoted as \textit{EOS}, is added to the character list and appended to the end of training target. 
During inference time, decoding ends once the \textit{EOS} is emitted. 
Apart from \textit{EOS}, $10$ Arabic digits and $52$ case-sensitive alphabets are predicted. 

This model is referred to as \textbf{rectification baseline} in our paper or \textbf{Rect} for short. These two names will be used interchangeably in the following sections.  

\subsubsection{Non-Rectification Baseline}
As shown in Fig.\ref{fig:pipeline}, we can also take off the rectification module and it becomes a variant of the convolutional recurrent neural network (CRNN)~\cite{shi2017end}, with RNN based transcription module stacked upon convolutional layers. 
The difference between CRNN and our model is that, the sequence output layer of CRNN is based on CTC, while our model uses the RNN-based encoder-decoder framework~\cite{sutskever2014sequence} which enables training and testing with variable lengths. 
We term our model without the rectification layer as \textbf{non-rectification baseline}, or \textbf{Non-Rect} for short. 
We only modify the rectification layer so that the Non-Rect model also serves as a benchmark for quantitative evaluation of the proposed techniques. 
The experiment results of Non-Rect can help attribute improvements to models and different techniques respectively. 
Also note that the input and output of the rectification module have the same size, and therefore, taking off it does not change the rest of the pipeline. 

\subsection{Training Settings} \label{bl:train}
\noindent \textbf{Training Target} Given input image $I\in R^{H\times W\times3}$ and ground truth text sequence $\{y_i\}^{T}_{t=1}$, the loss function is formulated as \textit{cross-entropy loss averaged over time}:

$$L=-\frac{1}{T}\sum^{T}_{t=1}log p(y_t|I)$$

Note that the last symbol of the ground truth sequence is always the special symbol \textit{EOS}. 

\noindent \textbf{Optimization} We use ADADELTA~\cite{zeiler2012adadelta} with default hyper-parameters (rho=9e-1, eps=1e-6, weight decay=0) to minimize the aforementioned loss function. 
Gradients are estimated using mini-batches with $512$ images that are randomly sampled from training set. 
We train $6$ epochs in total. 
We initialize the learning rate to $1.0$ for the first $4$ epochs, $0.1$ for the fifth epoch and $0.01$ for the sixth epoch. All experiments are performed on a Ubuntu machine with $4$ NVIDIA TITAN Xp graphics cards, each with 12GB memory. 

\subsection{Evaluation} \label{bl:eva}
We evaluate the models on a total of $9$ datasets. These datasets are briefly described here:

\noindent \textbf{IIIT 5K}~\cite{mishra2012scene} contains $3000$ horizontal testing images collected from the Internet that represent challenges in scene text recognition, such as the variety in font, size, and color. 

\noindent \textbf{Street View Text  (SVT)}~\cite{wang2011end} contains $647$ testing images collected from the Google Street View.  Many of them are corrupted by noise, blur and low resolution. 

\noindent \textbf{ICDAR 2003  (IC03)}~\cite{lucas2003icdar} recognition task contains $860$ horizontal focused images. 
Following previous works~\cite{wang2011end}, we only consider instances with at least $3$ characters. 

\noindent \textbf{ICDAR 2013  (IC13)}~\cite{karatzas2013icdar} inherit mostly from IC03 and contains $1015$ images. 

\noindent \textbf{ICDAR 2015 (IC15)}~\cite{karatzas2015icdar} is collected with Google Glasses without taking care of posing, position, and focus. 
The images are blurred, of low quality, and most of the text instances are oriented and small. 
We only consider instances with at least $3$ characters. 
The total number of images is $2077$. 

\noindent \textbf{SVT-Perspective (SVTP)}~\cite{quy2013recognizing} is collected in a similar way to SVT, with special focus on perspective text. 
It contains $645$ images. 

\noindent \textbf{CUTE 80}~\cite{risnumawan2014robust} is designed for curved text. 
It only contains $288$ instances for testing. 
By far, CUTE80 is the most widely used dataset to evaluate curved text recognition, but it has a small sized compared to other datasets. 

\noindent \textbf{Total-Text}~\cite{kheng2017total} is a dataset devoted to curved text detection and recognition. 
It has $1000$ scene images for training and  $500$ for testing. 
A large proportion of the instances are curved text and greatly oriented, making it difficult for both detection and recognition. 
We crop text instances from the test set, and obtain $2201$ cropped text images. 

\begin{figure}
\centering
\includegraphics[width=0.5\textwidth]{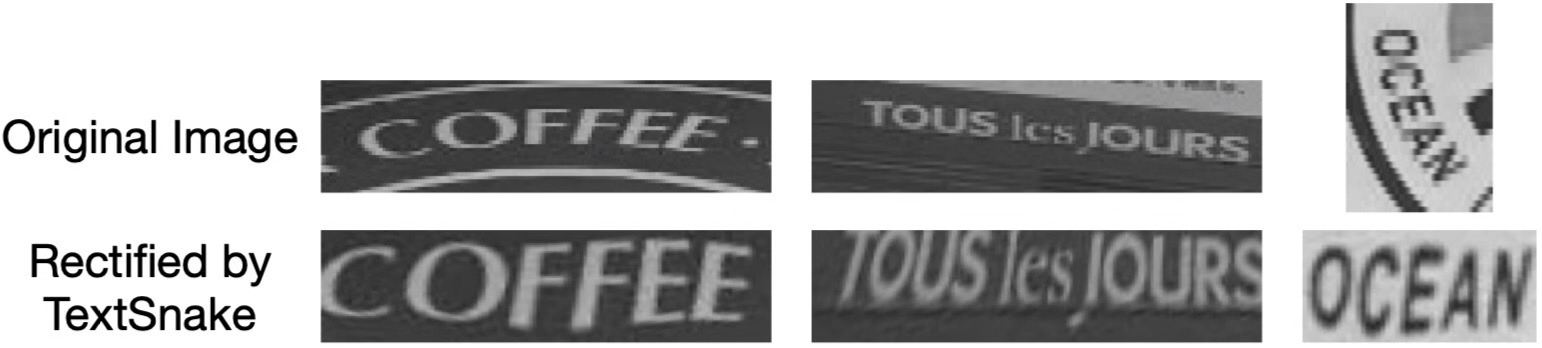}

\caption{Selected samples of RectTotal produced by applying TextSnake with ground truth geometry attributes. 
Note that, not only the text themselves are rectified, but the image background is also largely eliminated. }  
\label{fig:rect}
\vspace{-4mm}
\end{figure}

\noindent \textbf{Rectified Total-Text (RectTotal)} We propose \textbf{RectTotal} which is obtained by using the TextSnake~\cite{long2018textsnake} algorithm to rectify the test images using ground truth geometry attributes. 
If the detection module can precisely capture the shape of text instances instead of predicting them with groups of unordered pixels, curved text recognition algorithms may be less necessary. 
We include the results here for future reference. We also release RectTotal and the whole RectTotal dataset can also be found in our code repository. We select some samples for demonstration as shown in Fig.\ref{fig:rect}.

\noindent \textbf{ICDAR 2019 ArT (IC19-ArT)}\footnote{\url{https://rrc.cvc.uab.es/?ch=14}} is a competition devoted to curved text detection and recognition. 
The dataset for the recognition track contains $50K$ images for training and testing respectively, and contain both Latin script and Chinese Script. 
In this paper, we only consider Latin script.  The total number of Latin images in the training set is $36220$. 
Among these images, we randomly split $3000$ images as validation set, to match the size of other dataset. 
Also, as the test set is not available by the time this paper is written, we use this validation set as test set for evaluation. 

Among the $9$ dataset described above, the first $7$ datasets have been widely used in research. 
Therefore, we can compare the performance of our methods with previous ones. 
For the latter $2$ curved text dataset, we are the first to report experiment results. 
Without a reference point, we can only use them to evaluate the effectiveness of the proposed techniques in the form of ablation test. 
Nonetheless, they are much larger than the previous curved text baseline CUTE 80, and should act as better benchmarks. 
Therefore we release the results so that they can serve as a proper baseline for future research. 

\section{Select the Right Training Data} \label{data}

\subsection{Synthetic Text}

In the last few years, training solely on synthetic data has become a widely accepted norm. 
To be more precise, there are two synthetic datasets that are used by researchers: (1) the $800K$ synthetic images generated by the SynthText engine~\cite{gupta2016synthetic} that contain $7M$ cropped text images and (2) $9M$ cropped images from Synth90K~\cite{jaderberg2014synthetic}. 
Although SynthText has more realistic rendering, it is based on an imbalanced vocabulary. 
On the other hand, Synth90K has a balanced and large vocabulary. 
However, its images are monochrome, and the visual appearance is simpler than that of SynthText. 
Therefore, in most papers, the two datasets are combined to complement each other.  Such a norm continues even after the recognition of curved text became a hot topic. 

However, these two datasets only contain a low proportion of curved text, which may make it hard for recognizers to generalize to curved text. 
It is therefore reasonable and intuitive to generate synthetic curved text as a replacement. 
As SynthText provides an opensource code repository\footnote{\url{https://github.com/ankush-me/SynthText}} and Synth90K does not, we decide to work on SynthText. 

\subsubsection{Synthesizing curved text}
The SynthText engine renders text as a curve when the sampled text line contains only one word and the word contains no more than $10$ characters. 
Then, it samples a parabolic trajectory $y=\alpha x^2$ and places characters one by one symmetrically around the origin point. 
According to the code\footnote{\url{https://github.com/ankush-me/SynthText/blob/6640bf6a0d07b01cbb108984814af2aeb6e30344/text\_utils.py\#L59}}, $\alpha$ is uniformly sampled from $[-0.50, -0.45] \cup [0.50, 0.55]$. 
The potential problem is that, however, the engine's default parameters are set such that text source always attempts to sample a paragraph, i.e. multiple lines of text, and therefore it is only in rare condition that, there is only one word to render when the selected location of text is small enough. 
As a result, with the current parameters, curved text are rare in the original $800K$ dataset.

To solve this issue, we revise the text rendering module of the engine. 
The main idea of the new text rendering module is two-fold: (1) we increase the proportion of single word text so that more text are rendered as curved; (2) we randomly sample a radius and place text on the corresponding circle instead. 
The parameters are set such that there are a larger proportion of curved text and the curvature of text is greater. 
We also opensource the modified code for future research\footnote{\url{https://github.com/Jyouhou/CurvedSynthText}}. 
For technical details, readers an refer to the code.

\begin{figure}
\centering
\includegraphics[width=1\textwidth]{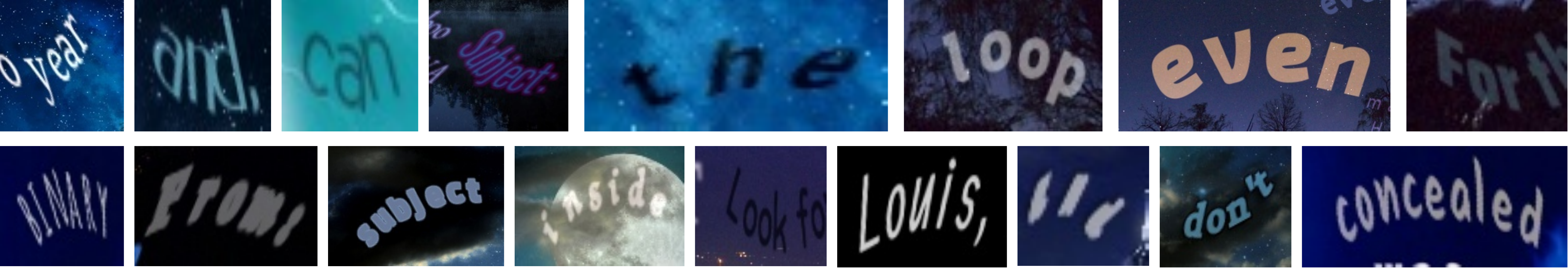}

\caption{Randomly selected samples generated by our modified SynthText Engine. }  
\label{fig:syn}
\end{figure}

We obtain $7M$ image crops in total. We estimate that around $10\%$ of the generated data are curved.  We randomly select some samples from our modified SynthText for visualization, as shown in Fig.\ref{fig:syn}. 
We refer to our version as \textit{CurvedSynth}. 

\subsubsection{Experiments with synthetic data}
Now we have three synthetic dataset: the original \textit{SynthText}, \textit{CurvedSynth}, 
and \textit{Synth90K}. 
To compare the effectiveness of different datasets, we first conduct an experiment with these datasets separately. 
When trained with \textit{Synth90K}, we transform the testing images to grey scale. 
We train the rectification baseline on these datasets respectively, and evaluate them on the $9$ real-world datasets. 
We also carry out the same experiment with the Non-Rect model to see how much datasets can benefit models. 

\begin{table}
\caption{Experiments with rectification baseline and non-rectification baseline trained on different synthetic data. For each dataset including RectTotal, we mark the highest score in bold. }
\label{tb-txp1}
\centering
{
\begin{tabular}{|c|c|c|c|c|c|c|c|c|c|c|}
\hline 
Model & Dataset & IIIT5K & SVT  & IC03 & IC13 & IC15 & SVTP & CUTE & \makecell{Total-Text\\(RectTotal)}  & IC19-ArT\tabularnewline
\hline
\hline
\multirow{3}{*}{\makecell{Non-Rect\\Baseline}} & \textit{CurvedSynth} & 0.906 & 0.836 & 0.927 & 0.888 & 0.698 & 0.695 & \textbf{0.847} & \makecell{0.692\\(0.627)} & 0.653 \tabularnewline
   & \textit{SynthText} & {0.912} & 0.849  & 0.929 & \textbf{0.915} & \textbf{0.731} & 0.722 & 0.799 & \makecell{0.629\\(\textbf{0.666})} & 0.649 \tabularnewline
& \textit{Synth90K} & 0.842 & 0.847 & 0.938 & 0.897 & 0.696 & 0.740 & 0.674 & \makecell{0.409\\({0.661})} & 0.568 \tabularnewline
\hline 
\multirow{3}{*}{\makecell{Rectification\\Baseline}} & \textit{CurvedSynth} & \textbf{0.920} & 0.850 & 0.922 & 0.893 & 0.722 & 0.740 & 0.837 & \makecell{\textbf{0.724} \\(0.660)} & \textbf{0.671} \tabularnewline
   & \textit{SynthText} & {0.916} & \textbf{0.853}  & 0.927 & 0.907 & {0.727} & 0.735 & 0.792 & \makecell{0.636 \\(0.663)} & 0.660 \tabularnewline
 & \textit{Synth90K} & {0.846} & {0.849} & \textbf{0.949} & 0.900 & 0.612 & \textbf{0.771} & 0.677 & \makecell{0.415 \\({{0.665}})} & 0.569 \tabularnewline
\hline 
\end{tabular}
}
\vspace{-2mm}
\end{table}

The experiment results are summarised in Tab.\ref{tb-txp1}. 
They show some consistency across models and dataset. 
We analyse these aspects, and inspect these evaluation datasets as well as model and training datasets one by one.  
We would present some very interesting findings. 

\noindent \textbf{Performance on straight text} We first focus on the $6$ horizontal datasets: III5K, SVT, IC03, IC13, IC15, and SVTP. 
On III5K and SVT, it is obvious that the rectification baseline performs better that the Non-Rect baseline. 
As for training datasets, models trained on \textit{CurvedSynth} parallels \textit{SynthText} overall, and \textit{Synth90K} ranks the worst. 
It is reasonable as IIIT5k actually contain a small proportion of curved and oriented text. SVT also has a few curved sample, and remember that, $1$ more correct image on SVT means an improvement by $0.15\%$. 
Therefore, the difference in the number of correctly recognized images may be little. 
We also note that, although models trained on \textit{SynthText} performs slightly better when using Non-Rect, \textit{CurvedSynth} soon catches up when using rectification layer. 
It is reasonable to attribute it to the compatibility between rectification layer and curved text. 
Other wise, the curved text are distorted greatly when resized to $64\times256$, which makes it difficult to recognize. 

\begin{figure}
\centering
\includegraphics[width=1\textwidth]{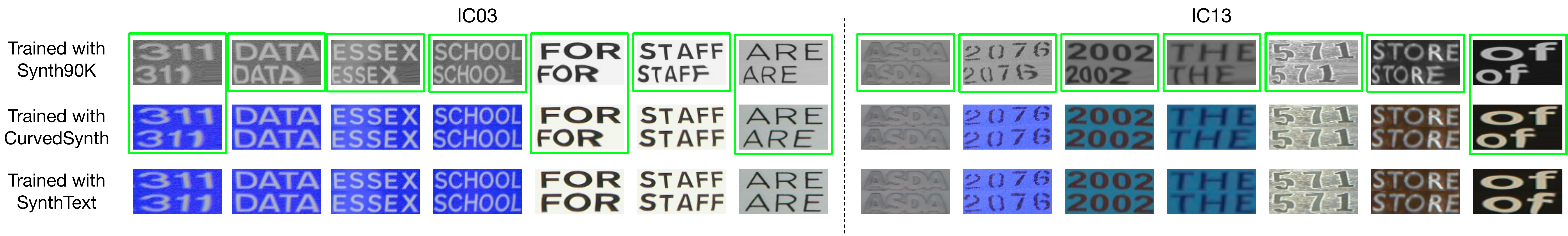}

\caption{Randomly selected results by rectification baseline from IC03 and IC13 trained with different training datasets. The upper part of each sample is the input images which are resized to $64\times256$ and the bottom is the rectified images.}  
\label{fig:exp1-hori}
\end{figure}

On IC03 and IC13, things are different. As IC03 and IC15 contain few curved text, it is as expected that rectification layer brings no improvements. 
Besides, the fact that models trained with \textit{CurvedSynth} performs worse than \textit{SynthText} may indicate that redundant curved training examples may make the model less competent on purely horizontal text. 
It is worth-noting that, models trained on \textit{Synth90K} benefits from the rectification layer on IC03 by a considerable margin. According to a recent survey~\cite{long2018scene}, the state-of-the-art performance on IC03 is $94.5\%$ when trained on \textit{Synth90K}+\textit{SynthText}, which is surpassed by ours. 
To understand what contributes to it, we inspect the evaluation results by checking each testing images one-by-by manually. We notice a outstanding phenomenon, select some representatives, and show them in Fig.\ref{fig:exp1-hori}. 
We notice that the rectification layer actually shorten some text so that the aspect ratios of characters look better, although leaving some blank at the end. 
Such resizing is frequently found in results produced by models that are trained on \textit{Synth90K}. 
This phenomenon also exists in models trained with \textit{CurvedSynth}, but not in the original \textit{SynthText}. 
It is also not found in IC13, while there is also little improvement on IC13. 
We assume that encouraging such phenomenon can be an important technical trick. \label{ShrankText}

On IC15 and SVTP, we carry out similar in-depth inspection on testing images one-by-one and found similar phenomenon. Models trained with \textit{Synth90K} seems to be better at resizing the text to a more normal aspect ratio. 
It remains a challenge to investigate the reason behind this. 
We suppose it is because \textit{Synth90K} has a balanced vocabulary, and therefore has more long words, while the SynthText engine uses an imbalanced vocabulary, and in natural corpus, most high-frequency words are short. 
Thus, models trained on Synth90K would have to learn to resize characters so that words with varying lengths can fit the input size better. 

\noindent \textbf{Performance on curved text} On CUTE80, Total-Text, and IC19-Art, we observe consistent improvements brought by our \textit{CurvedSynth} over the original \textit{SynthText}. It results in an improvement of $4$-$5\%$ on CUTE80, around $6$-$9\%$ on Total-Text, and about $1\%$ on IC19-ArT, depending on the model we use. 
Especially, when we use rectification baseline, the use of \textit{CurvedSynth} results in greater improvements over models trained with \textit{SynthText} on Total-Text, from $6.3\%$ to $8.8\%$,  and IC19-ArT, from $0.4\%$ to $1.1\%$. 
Although performance on CUTE80 drops after adding rectification layer, it can be attributed to the small size of CUTE80 and the variance it may have.

On the other hand, when trained on \textit{SynthText} or \textit{Synth90K} which contain few curved text, the rectification layer brings only insignificant improvements. The improvements are less than $0.3\%$ on CUTE80, $0.6\%$ on Total-Text, and $0.1$-$1.0\%$ on IC19-ArT. These numbers are quite small compared to improvements by introducing our \textit{CurvedSynth}. When using \textit{CurvedSynth}, the rectification layer can boost performance by $3.2\%$ on Total-Text and $1.8\%$ on IC19-ArT. We can draw a conclusion that, our proposed \textit{CurvedSynth} can help rectification layer reach its potential by providing appropriate data. 

\noindent \textbf{Performance on rectified  Total-Text} We also evaluate all models on the RectTotal dataset we create using the TextSnake algorithm as specified in Section \ref{bl:eva}. 
The results are marked by a set of brackets in the same table. 
We notice that the Non-Rect baseline trained on \textit{SynthText} with few curved text instances achieve the highest score of $66.6\%$. 
The performance of models except Non-Rect trained on \textit{CurvedSynth} is not significantly different from this best model. 
The reason why Non-Rect trained on \textit{CurvedSynth} achieves the lowest accuracy may be attributed to the fact that, since the images in RectTotal are rectified, background is eliminated and the text compactly fit into and fill the images, as shown in the bottom row of Fig.\ref{fig:rect}. 
This, as it turns out, makes these images visually more similar to the training images in \textit{SynthText} and \textit{Synth90K}. 
On the contrary, the curved text in \textit{CurvedSynth} contain background and complex spatial alignment, which is actually out-of-distribution for RectTotal. 

Another counter-intuitive phenomenon is that, models trained on \textit{CurvedSynth} perform worse on RectTotal overall than on the original Total-Text. 
Note that, as the ground-truth polygon annotations of Total-Text is imprecise, the rectified images may be warped. 
Therefore, the performance gap may be due to the extra distortion brought up by this process. 
Basically, experiments on RectTotal lead us to the conclusion that, with a good detection model that can capture the shape of text and further rectify it, sophisticated model designs and tricks may be unnecessary. 
Specifically, we can observe absolute improvements in accuracy by $25\%$ on Total-Text for models trained on \textit{Synth90K} if we rectify them in advance. 
But for now, we still need them as IC19-ArT is not rectified. 

\subsubsection{Experiments with multiple synthetic data} 

\begin{table}
\caption{Experiments with our rectification baseline trained on multiple synthetic datasets compared with previous state-of-the-art methods. Performance scores marked with `-` indicate that they are not reported. `*` indicates results provided by asking the authors but not presented in the paper. 
Again, performance on RectTotal is included only for future reference. 
Best results are marked in \textbf{bold}. 
All scores here are obtained by training models solely on synthetic data. 
}
\label{tb-joint}
\centering
{
\small
\begin{tabular}{|c|c|c|c|c|c|c|c|c|c|c|}
\hline 
Model & Dataset & IIIT5K & SVT  & IC03 & IC13 & IC15 & SVTP & CUTE & \makecell{Total-Text\\(RectTotal)}  & IC19-ArT\tabularnewline
\hline
\hline
SAR\cite{li2018show} & \makecell{\textit{SynthText}\\ + \textit{Synth90K}} & {0.915} & {0.845} & {-} & 0.910 & 0.692 & {0.764} & 0.833 & - & -\tabularnewline
\hline 
CA-FCN\cite{liao2018scene} & \makecell{\textit{SynthText}\\ + private data} & {0.919} & {0.864} & {-} & 0.915 & - & {-} & 0.799 & \makecell{0.616$^*$ \\({-})} & -\tabularnewline
\hline 
{\makecell{Rectification\\Baseline}} & \makecell{\textit{CurvedSynth}\\ + \textit{Synth90K}} & {\textbf{0.948}} & {\textbf{0.896}} & {{0.958}} & \textbf{0.928} & \textbf{0.782} & {\textbf{0.816}} & \textbf{0.896} & \makecell{\textbf{0.763} \\({\textbf{0.734}})} & {0.721} \tabularnewline
\hline 
\end{tabular}
}
\vspace{-2mm}
\end{table}

As most recent methods are trained on \textit{Synth90K} and \textit{SynthText} jointly, we also present our results as trained on \textit{Synth90K} and \textit{CurvedSynth} jointly, which is shown in Tab.\ref{tb-joint}. 
The total number of training data is $16M$. 
With our \textit{CurvedSynth}, we surpass recent baselines on all datasets by a large margin. We achieve significant improvements on CUTE80 by $6.3\%$ and on Total-Text by $14.7\%$. The comparison verifies the effectiveness of our \textit{CurvedSynth} dataset. 

\subsection{Mixing with Real World Data}

As there are several annotated real world datasets, it seems a potential way to boost performance using read world images. 
However, the quantity of real world images is much smaller than synthetic data. 
It is then valuable to find out an optimal sampling scheme to balance real world data and synthetic data. 
This section is dedicated to this target. 

\begin{table}
\caption{Sizes of real world datasets with annotations. The last column entitled \textit{sum} is the total amount of data for train/test respectively. }
\label{tb-data}
\centering
{
\begin{tabular}{|c|c|c|c|c|c|c|c|c|c|c|c|}
\hline 
  & IIIT5K & SVT & IC03 & IC13 & IC15 & SVTP & CUTE & Total-Text & IC19-ArT & COCO & sum \tabularnewline
\hline
\hline
train & 2000 & 257 & 1156 & 848 & 4467 & 0  & 0 & 9267 & 33220 & 13943 & 65158 \tabularnewline
\hline 
test & 3000 & 647 & 860 & 1015 & 1811 & 645 & 288 & 2201 & 3000 & 0 & 13467\tabularnewline
\hline 
\end{tabular}
}
\vspace{-2mm}
\end{table}

We collect available real world data and summarize them in Tab.\ref{tb-data}. 
As IC19-ArT is only released recently, we do not add it into our training set for fair comparison with previous works that use real world data~\cite{li2018show}. 
Together we obtain $31938$ real world images. 
Since the number of synthetic data is much larger than that of real world images, we give real world data a larger sampling weight. 
We carry out experiments with different sampling weights such that the real world data account for $5\%$, $10\%$, $15\%$, $20\%$ and $25\%$ of the total training data. Also note that, we temporarily treat all real datasets the same and ignore the differences in their sizes. 

\begin{table}
\caption{Experiments with synthetic data mixed with real world data. 
}
\label{tb-real}
\centering
{
\small
\begin{tabular}{|c|c|c|c|c|c|c|c|c|c|c|c|}
\hline 
Model & \makecell{Real World \\ Data Ratio ($\%$)} & IIIT5K & SVT  & IC03 & IC13 & IC15 & SVTP & CUTE & Total-Text  & IC19-ArT&Avg\tabularnewline
\hline
\hline
SAR\cite{li2018show} & $13.5$ & {0.950} & {0.912} & {-} & 0.940 & 0.788 & {\textbf{0.864}} & 0.896 & - & - & - \tabularnewline
\hline 
\hline 
{\makecell{Rectification\\Baseline}} & 0 & {0.948} & {0.896} & {0.958} & 0.928 & 0.782 & {0.816} & 0.896 & 0.763 & 0.721 & 0.834 \tabularnewline
\hline 
{\makecell{Rectification\\Baseline}} & 5 & {\textbf{0.961}} & {\textbf{0.927}} & {0.952} & 0.949 & 0.834 & {0.842} & 0.938 & \textbf{0.817} & {0.768} & {0.868} \tabularnewline
\hline 
{\makecell{Rectification\\Baseline}} & 10 & {0.958} & {0.924} & {{0.964}} & \textbf{0.953} & 0.832 & {{0.851}} & 0.931 & 0.814 & {0.762} & {{0.866}} \tabularnewline
\hline 
{\makecell{Rectification\\Baseline}} & 15 & {0.957} & {0.921} & {0.955} & 0.945 & \textbf{0.847} & {0.842} & 0.934 & 0.815 & \textbf{0.772} & \textbf{0.869} \tabularnewline
\hline 
{\makecell{Rectification\\Baseline}} & 20 & {{0.956}} & {0.926} & {\textbf{0.966}} & 0.948 & 0.839 & {0.840} & 0.938 & 0.807 & {0.759} & 0.864 \tabularnewline
\hline 
{\makecell{Rectification\\Baseline}} & 25 & {0.954} & {0.924} & {0.958} & 0.951 & {0.838} & {0.828} & \textbf{0.951} & 0.809 & 0.752 & 0.862 \tabularnewline
\hline 
\end{tabular}
}
\vspace{-2mm}
\end{table}

As shown in Tab.\ref{tb-real}, our model outperforms previous methods that use real world data by a large margin consistently for different real world data ratios. 
We achieve improvements on all datasets except SVTP. Notably, IC15, SVTP, CUTE80, Total-Text and IC19-Art all benefit significantly from real world data. 

For an overall comparison, we compute the micro average score by taking the weighted average of all accuracy with dataset sizes as weights, and the average scores are shown in the last column labeled as \textit{Avg}. 
In this sense, the rectification baseline performs best when the real world data should take up $15\%$ of the total training data. 

As for performance on curved text datasets, a ratio of $10\%$ and $15\%$ result in fairly comparative results. A proportion of $15\%$ for real world data results in slightly higher accuracy in Total-Text and a $1.0\%$ improvement is insignificant. Since the ratio is not making very large difference on most datasets, we would continue with $15\%$ in later experiments on IC19-ArT due to its superiority both in curved text and in the overall way. 

\section{Model Design Tweaks} \label{tweak}
The previous sections mainly focus on the selection of data. 
In this part, we mainly talk about some model modifications that improve the performance of our rectification based methods. 
For fair comparison with most previous works as well as previous sections, we only train our method on synthetic data, i.e. \textit{Synth90K} + \textit{CurvedSynth}. 

\subsection{Keeping Aspect-Ratio for Input Images}
As suggested in Section \ref{ShrankText}, the rectification layer learns to resizing the input images, and it is likely to be the reason why it performs better on those datasets. 
We suppose an aspect-ratio keeping input would benefit recognizers. 
Therefore, we design a new and intuitive pre-processing step as specified here. 

\begin{figure}
\centering
\includegraphics[width=1.0\textwidth]{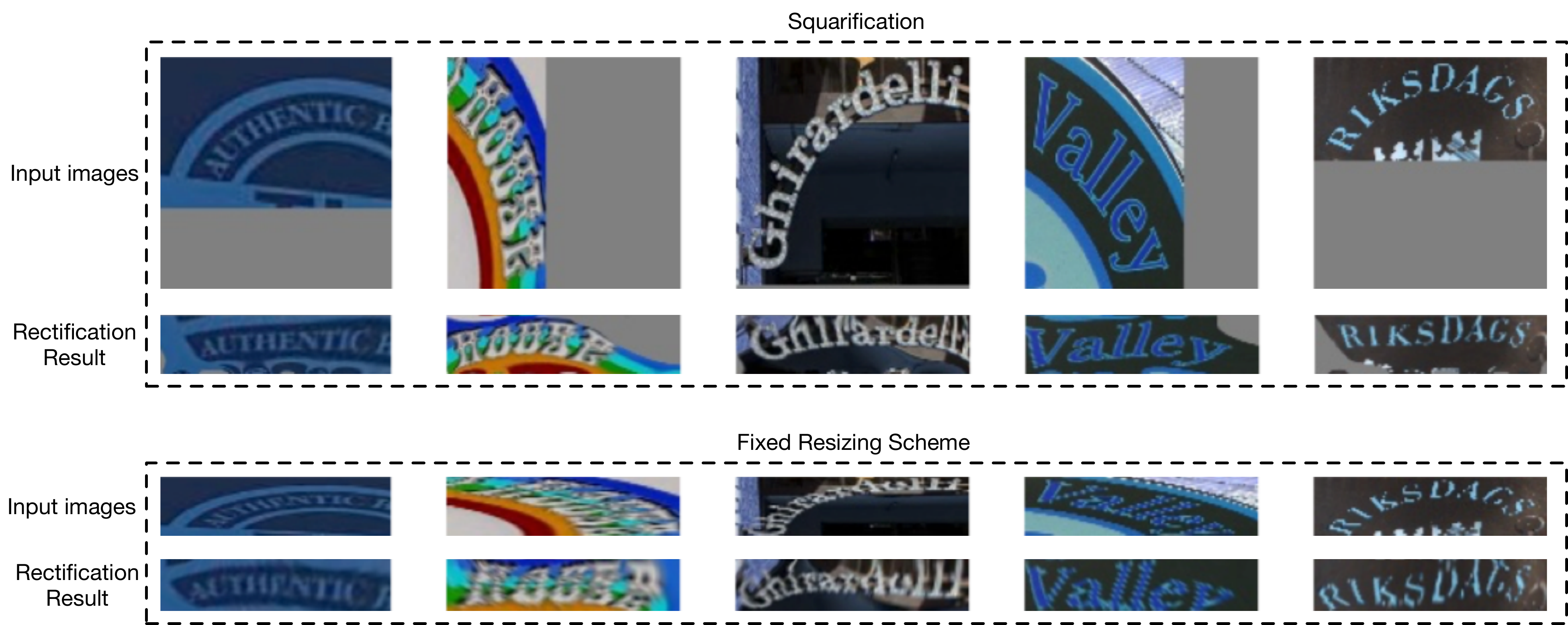}

\caption{Visualization of rectification results for squarization compared with fixed resizing scheme. 
Within each dotted frame, in the upper part lies the input image, and in bottom part lies the rectification results. 
We can see that, rectification based on square images produce higher image quality, more clear rectified images and better overall rectification. }  
\label{fig:squa}
\vspace{-4mm}
\end{figure}

For input images with different sizes, we first resize the long sides to $256$ pixels, and short sides resized accordingly to keep the original aspect-ratios. 
Then, we pad \textit{grey} borders to the short sides such that the pre-processed images are square. 
We select and visualize some samples in Fig.\ref{fig:squa}. It is obvious that our pre-processing is more friendly to curved text. 
Compared with fixed-size scheme that resizes images to $64\times256$ uniformly, our method keeps the images semantically legible. We term our pre-processing as \textbf{squarization}.

\subsubsection{Random Rotation} 
Another advantage of squarization is that, now images can be rotated freely in both training and testing stages. 
As instances in curved text dataset usually have oriented or even vertical arrangement, it is important to incorporate such situations in the training stages as well. Therefore, we propose to randomly rotate the images during training. Specifically, the images are randomly rotated $90$, $180$, or $270$ degrees with a probability of $5\%$ respectively. 

\subsection{Rectification on Images}
While rectification layer can alleviate distortions to some extent, it only operates on the extracted feature maps and therefore the features may still be affected by the distortion in images. 
Therefore, we carry out experiments to inspect whether performing rectification on images can benefit recognition. 

Specifically, the images are first fed forward to obtain the control points as our rectification baseline does. Then the control points are used to rectify the input images. 
Then the rectified images are again fed into the same network. 
The recognition module is applied to the feature maps extracted from the rectified images. 

In implementation, such two-pass computation increases memory usage by nearly one time, and makes it impossible to fit in the GPUs we have. Therefore, in the first pass, we downsample the input images to half size using bilinear interpolation, but the second pass is performed on the images of original resolution. By reducing the input size of the first pass, the rectification module based on fully connected networks can also remain unchanged. 

\subsection{Experiments} 

\begin{table}
\caption{Results of experiments with model modifications. 
}
\label{tb-tricks}
\centering
{
\small
\begin{tabular}{|c|c|c|c|c|c|c|c|c|c|c|}
\hline 
Model & Modification & IIIT5K & SVT  & IC03 & IC13 & IC15 & SVTP & CUTE & Total-Text  & IC19-ArT\tabularnewline
\hline
\hline
{\makecell{Rectification\\Baseline}} & None & {0.948} & {0.896} & {\textbf{0.958}} & \textbf{0.928} & 0.782 & {0.816} & 0.896 & 0.763 & 0.721 \tabularnewline
\hline 
{\makecell{Rectification\\Baseline}} & Squarization & {\textbf{0.950}} & {{0.898}} & {0.950} & {0.927} & 0.782 & {0.809} & 0.892 & 0.769 & 0.724 \tabularnewline
\hline 
{\makecell{Rectification\\Baseline}} & \makecell{Squarization\\ + Random Rotation} & 0.949 & {0.898} & {{{0.948}}} & {0.925} & {0.781} & {{\textbf{0.817}}} & 0.889 & \textbf{0.781} & \textbf{0.728} \tabularnewline
\hline 
{\makecell{Rectification\\Baseline}} & \makecell{Squarization\\ + Rectify Image} & 0.946 & {\textbf{0.904}} & {0.953} & 0.925 & \textbf{0.792} & {0.795} & 0.800 & 0.776 & {0.723}  \tabularnewline
\hline 
{\makecell{Rectification\\Baseline}} & All & {{0.942}} & {0.881} & {0.948} & 0.920 & 0.771 & {0.803} & \textbf{0.903} & 0.761 & 0.720  \tabularnewline
\hline 
\end{tabular}
}
\vspace{-2mm}
\end{table}

We list the experiment results in Tab.\ref{tb-tricks}. 
We can see that the combination of squarization and random rotation boost performance on curved text recognition, with improvements on Total-Text by $1.8\%$, and IC19-ArT by $0.7\%$. 
Besides, squarization itself only results in slight improvements on nearly all datasets. 
However, we argue that squarization is still important because it enables the random rotation trick that can bring more improvements. 

Unfortunately, the practice of rectifying images does not seem to be very effective. Although it scores highest on SVT with an improvement by $0.8\%$ and IC15 by $1.0\%$, it fails on other datasets. We suppose rectification on images may make it difficult to train.  

\section{Evaluation on ICDAR-2019} \label{ic19}

In this part, we reveal details of how we build our recognition system for participation in IC19-ArT. 

\noindent \textbf{Data} We pool all available data together: synthetic data including \textit{CurvedSynth} and \textit{Synth90K}, all real world data as listed in Tab.\ref{tb-data} except the $3000$ validation images we sample from IC19-ArT. We have $16M$ synthetic data in total, and $75625$ real world images. During training, we sample real world data with a larger weight so that real world data take up $15\%$ of total training data. We use the $3000$ validation set to select the best models. 

\noindent \textbf{Model} We train an ensemble of the following models on the aforementioned data: (1) rectification baseline; (2) squarified rectification model with random rotation; (3) rectification baseline with ResNet152; (4) rectification baseline with rectification on images. We use the validation set to select the best model parameters for each of the $4$ models and combine the prediction results via a simple voting mechanism: we simply select the text that appear most from the $4$ predictions. 

The $4$ models achieve validation accuracy from $76.5\%$ and $78.5\%$. The final accuracy on the test set that is not yet published is $74.3\%$.

\section{Conclusion} \label{conc}
In this paper, we investigate several aspects that may affect the training and final performance of rectification based scene text recognizers on curved text datasets. 
We implement a rectification-based method with a standard ResNet+FPN as feature extractor, a rectification module, and a vanilla attentional RNN based encoder-decoder sequence learning module. 
We conduct comprehensive experiments to study the task of curved text recognition, and we finally arrive at the following conclusions:

\begin{enumerate}
    \item \textbf{Training data that contain curved text are very important yet economical.} Despite the latest fashion of training solely on the standard SynthText and Synth90K for fair comparison, these two datasets contain few curved instances, and are therefore not suitable for curved text recognition. Algorithms trained on these datasets are not able to reach their full potential. We demonstrate this by training a previous state-of-the-art method on curved synthetic text, and boost its performance by as much as $6.3\%$ on CUTE80, even surpassing the most recent state-of-the-art trained on both synthetic data and real world data. On the other hand, the synthesis of curved text is effortless and economical, and can replace existing datasets in a drop-in fashion. The proposed \textit{CurvedSynth} is a better dataset to study the task of curved text recognition.
    \item \textbf{Mixing synthetic data with real ones also makes a difference.} This rule of thumb is self-evident, but technical details are important. As shown in our experiments, it is important to find an appropriate proportion for the real world data. Besides, these datasets vary greatly in size, from a few hundreds to tens of thousands. Thus, it is also valuable to design a method to find an optimal ratio among real world datasets.
    \item \textbf{Squarization works, and brings up new possibility.} Another standard norm we challenges is the fixed-size  pre-processing that resizes all input images to the same preset size, which is usually $64\times256$. Although the proposed squarization only results in slight improvements on curved datasets and only parallel results on straight text, it enables the use of random rotation, which can boost the performance on curved text to some extent. For future directions, it is also worthwhile to study inputs with variable lengths/widths, as done by the NLP community. This may be important for long text, such as Chinese, Japanese, Korean, and etc.. 
    \item \textbf{Recognizers are more robust to shape when working on images rectified by select detectors. } We propose \textbf{RectTotal}, a dataset obtained by applying TextSnake to rectify test images of Total-Text using the ground truth polygon-based labels. Experiment results show that, even if algorithms are solely trained on straight text and not equipped with special mechanism for curved text recognition, they can achieve equally good results if the text are already rectified. This result places more importance on detectors that can capture the shape of text, such as TextSnake and CRAFT~\cite{baek2019character}. Text detection and recognition should be co-designed to achieve better results. 
\end{enumerate}

Finally, we release the training details for our participation in IC19-ArT. We hope it will help researchers in designing better algorithms and further mining new challenges that are not yet noticed by the community. 

\bibliography{references}

\begin{thebibliography}{10}

\bibitem{baek2019character}
Youngmin Baek, Bado Lee, Dongyoon Han, Sangdoo Yun, and Hwalsuk Lee.
\newblock Character region awareness for text detection.
\newblock In {\em Proceedings of the IEEE Conference on Computer Vision and
  Pattern Recognition (CVPR)}, pages 9365--9374, 2019.

\bibitem{cheng2017focusing}
Zhanzhan Cheng, Fan Bai, Yunlu Xu, Gang Zheng, Shiliang Pu, and Shuigeng Zhou.
\newblock Focusing attention: Towards accurate text recognition in natural
  images.
\newblock In {\em 2017 IEEE International Conference on Computer Vision
  (ICCV)}, pages 5086--5094. IEEE, 2017.

\bibitem{kheng2017total}
Chee~Kheng Ch'ng and Chee~Seng Chan.
\newblock Total-text: A comprehensive dataset for scene text detection and
  recognition.
\newblock In {\em Document Analysis and Recognition (ICDAR), 2017 14th IAPR
  International Conference on}, volume~1, pages 935--942. IEEE, 2017.

\bibitem{Deng2018}
Deng Dan, Liu Haifeng, Li~Xuelong, and Cai Deng.
\newblock Pixellink: Detecting scene text via instance segmentation.
\newblock In {\em Proceedings of AAAI, 2018}, 2018.

\bibitem{gao2017reading}
Yunze Gao, Yingying Chen, Jinqiao Wang, and Hanqing Lu.
\newblock Reading scene text with attention convolutional sequence modeling.
\newblock {\em arXiv preprint arXiv:1709.04303}, 2017.

\bibitem{gupta2016synthetic}
Ankush Gupta, Andrea Vedaldi, and Andrew Zisserman.
\newblock Synthetic data for text localisation in natural images.
\newblock In {\em Proceedings of the IEEE Conference on Computer Vision and
  Pattern Recognition (CVPR)}, pages 2315--2324, 2016.

\bibitem{He_2017_Res}
Kaiming He, Xiangyu Zhang, Shaoqing Ren, and Jian Sun.
\newblock Deep residual learning for image recognition.
\newblock In {\em In Proceedings of the IEEE conference on computer vision and
  pattern recognition (CVPR)}, 2016.

\bibitem{hochreiter1997long}
Sepp Hochreiter and J{\"u}rgen Schmidhuber.
\newblock Long short-term memory.
\newblock {\em Neural computation}, 9(8):1735--1780, 1997.

\bibitem{jaderberg2014synthetic}
Max Jaderberg, Karen Simonyan, Andrea Vedaldi, and Andrew Zisserman.
\newblock Synthetic data and artificial neural networks for natural scene text
  recognition.
\newblock {\em arXiv preprint arXiv:1406.2227}, 2014.

\bibitem{jaderberg2015spatial}
Max Jaderberg, Karen Simonyan, Andrew Zisserman, et~al.
\newblock Spatial transformer networks.
\newblock In {\em Advances in neural information processing systems}, pages
  2017--2025, 2015.

\bibitem{karatzas2015icdar}
Dimosthenis Karatzas, Lluis Gomez-Bigorda, Anguelos Nicolaou, Suman Ghosh,
  Andrew Bagdanov, Masakazu Iwamura, Jiri Matas, Lukas Neumann,
  Vijay~Ramaseshan Chandrasekhar, Shijian Lu, et~al.
\newblock Icdar 2015 competition on robust reading.
\newblock In {\em Document Analysis and Recognition (ICDAR), 2015 13th
  International Conference on}, pages 1156--1160. IEEE, 2015.

\bibitem{karatzas2013icdar}
Dimosthenis Karatzas, Faisal Shafait, Seiichi Uchida, Masakazu Iwamura,
  Lluis~Gomez i~Bigorda, Sergi~Robles Mestre, Joan Mas, David~Fernandez Mota,
  Jon~Almazan Almazan, and Lluis~Pere de~las Heras.
\newblock Icdar 2013 robust reading competition.
\newblock In {\em Document Analysis and Recognition (ICDAR), 2013 12th
  International Conference on}, pages 1484--1493. IEEE, 2013.

\bibitem{lee2016recursive}
Chen-Yu Lee and Simon Osindero.
\newblock Recursive recurrent nets with attention modeling for ocr in the wild.
\newblock In {\em Proceedings of the IEEE Conference on Computer Vision and
  Pattern Recognition (CVPR)}, pages 2231--2239, 2016.

\bibitem{li2018show}
Hui Li, Peng Wang, Chunhua Shen, and Guyu Zhang.
\newblock Show, attend and read: A simple and strong baseline for irregular
  text recognition.
\newblock {\em AAAI}, 2019.

\bibitem{liao2019synthtext3d}
Minghui Liao, Boyu Song, Minghang He, Shangbang Long, Cong Yao, and Xiang Bai.
\newblock Synthtext3d: Synthesizing scene text images from 3d virtual worlds.
\newblock {\em arXiv preprint arXiv:1907.06007}, 2019.

\bibitem{liao2018scene}
Minghui Liao, Jian Zhang, Zhaoyi Wan, Fengming Xie, Jiajun Liang, Pengyuan Lyu,
  Cong Yao, and Xiang Bai.
\newblock Scene text recognition from two-dimensional perspective.
\newblock {\em AAAI}, 2019.

\bibitem{Lin_2017_CVPR}
Tsung-Yi Lin, Piotr Dollar, Ross Girshick, Kaiming He, Bharath Hariharan, and
  Serge Belongie.
\newblock Feature pyramid networks for object detection.
\newblock In {\em The IEEE Conference on Computer Vision and Pattern
  Recognition (CVPR)}, July 2017.

\bibitem{liu2016star}
Wei Liu, Chaofeng Chen, Kwan-Yee~K Wong, Zhizhong Su, and Junyu Han.
\newblock Star-net: A spatial attention residue network for scene text
  recognition.
\newblock In {\em BMVC}, volume~2, page~7, 2016.

\bibitem{long2015fully}
Jonathan Long, Evan Shelhamer, and Trevor Darrell.
\newblock Fully convolutional networks for semantic segmentation.
\newblock In {\em Proceedings of the IEEE Conference on Computer Vision and
  Pattern Recognition (CVPR)}, pages 3431--3440, 2015.

\bibitem{long2018scene}
Shangbang Long, Xin He, and Cong Ya.
\newblock Scene text detection and recognition: The deep learning era.
\newblock {\em arXiv preprint arXiv:1811.04256}, 2018.

\bibitem{long2018textsnake}
Shangbang Long, Jiaqiang Ruan, Wenjie Zhang, Xin He, Wenhao Wu, and Cong Yao.
\newblock Textsnake: A flexible representation for detecting text of arbitrary
  shapes.
\newblock In {\em In Proceedings of European Conference on Computer Vision
  (ECCV)}, 2018.

\bibitem{lucas2003icdar}
Simon~M Lucas, Alex Panaretos, Luis Sosa, Anthony Tang, Shirley Wong, and
  Robert Young.
\newblock Icdar 2003 robust reading competitions.
\newblock In {\em null}, page 682. IEEE, 2003.

\bibitem{mishra2012scene}
Anand Mishra, Karteek Alahari, and CV~Jawahar.
\newblock Scene text recognition using higher order language priors.
\newblock In {\em BMVC-British Machine Vision Conference}. BMVA, 2012.

\bibitem{quy2013recognizing}
Trung Quy~Phan, Palaiahnakote Shivakumara, Shangxuan Tian, and Chew Lim~Tan.
\newblock Recognizing text with perspective distortion in natural scenes.
\newblock In {\em Proceedings of the IEEE International Conference on Computer
  Vision (ICCV)}, pages 569--576, 2013.

\bibitem{risnumawan2014robust}
Anhar Risnumawan, Palaiahankote Shivakumara, Chee~Seng Chan, and Chew~Lim Tan.
\newblock A robust arbitrary text detection system for natural scene images.
\newblock {\em Expert Systems with Applications}, 41(18):8027--8048, 2014.

\bibitem{Shi_2017_CVPR}
Baoguang Shi, Xiang Bai, and Serge Belongie.
\newblock Detecting oriented text in natural images by linking segments.
\newblock In {\em The IEEE Conference on Computer Vision and Pattern
  Recognition (CVPR)}, July 2017.

\bibitem{shi2017end}
Baoguang Shi, Xiang Bai, and Cong Yao.
\newblock An end-to-end trainable neural network for image-based sequence
  recognition and its application to scene text recognition.
\newblock {\em IEEE transactions on pattern analysis and machine intelligence},
  39(11):2298--2304, 2017.

\bibitem{shi2016robust}
Baoguang Shi, Xinggang Wang, Pengyuan Lyu, Cong Yao, and Xiang Bai.
\newblock Robust scene text recognition with automatic rectification.
\newblock In {\em Proceedings of the IEEE Conference on Computer Vision and
  Pattern Recognition (CVPR)}, pages 4168--4176, 2016.

\bibitem{shi2018aster}
Baoguang Shi, Mingkun Yang, XingGang Wang, Pengyuan Lyu, Xiang Bai, and Cong
  Yao.
\newblock Aster: An attentional scene text recognizer with flexible
  rectification.
\newblock {\em IEEE transactions on pattern analysis and machine intelligence},
  31(11):855--868, 2018.

\bibitem{su2014accurate}
Bolan Su and Shijian Lu.
\newblock Accurate scene text recognition based on recurrent neural network.
\newblock In {\em Asian Conference on Computer Vision}, pages 35--48. Springer,
  2014.

\bibitem{sutskever2014sequence}
Ilya Sutskever, Oriol Vinyals, and Quoc~V Le.
\newblock Sequence to sequence learning with neural networks.
\newblock In {\em Advances in neural information processing systems}, pages
  3104--3112, 2014.

\bibitem{tian2019learning}
Zhuotao Tian, Michelle Shu, Pengyuan Lyu, Ruiyu Li, Chao Zhou, Xiaoyong Shen,
  and Jiaya Jia.
\newblock Learning shape-aware embedding for scene text detection.
\newblock In {\em Proceedings of the IEEE Conference on Computer Vision and
  Pattern Recognition}, pages 4234--4243, 2019.

\bibitem{wang2011end}
Kai Wang, Boris Babenko, and Serge Belongie.
\newblock End-to-end scene text recognition.
\newblock In {\em Computer Vision (ICCV), 2011 IEEE International Conference
  on}, pages 1457--1464. IEEE, 2011.

\bibitem{warps1989thin}
Fred L Bookstein~Principal Warps.
\newblock Thin-plate splines and the decompositions of deformations.
\newblock {\em IEEE Transactions on Pattern Analysis and Machine Intelligence},
  11(6), 1989.

\bibitem{xu2015show}
Kelvin Xu, Jimmy Ba, Ryan Kiros, Kyunghyun Cho, Aaron Courville, Ruslan
  Salakhudinov, Rich Zemel, and Yoshua Bengio.
\newblock Show, attend and tell: Neural image caption generation with visual
  attention.
\newblock In {\em International Conference on Machine Learning}, pages
  2048--2057, 2015.

\bibitem{yin2017scene}
Fei Yin, Yi-Chao Wu, Xu-Yao Zhang, and Cheng-Lin Liu.
\newblock Scene text recognition with sliding convolutional character models.
\newblock {\em arXiv preprint arXiv:1709.01727}, 2017.

\bibitem{zeiler2012adadelta}
Matthew~D Zeiler.
\newblock Adadelta: an adaptive learning rate method.
\newblock {\em arXiv preprint arXiv:1212.5701}, 2012.

\bibitem{zhan2018verisimilar}
Fangneng Zhan, Shijian Lu, and Chuhui Xue.
\newblock Verisimilar image synthesis for accurate detection and recognition of
  texts in scenes.
\newblock 2018.

\bibitem{Zhou_2017_CVPR}
Xinyu Zhou, Cong Yao, He~Wen, Yuzhi Wang, Shuchang Zhou, Weiran He, and Jiajun
  Liang.
\newblock {EAST}: An efficient and accurate scene text detector.
\newblock In {\em The IEEE Conference on Computer Vision and Pattern
  Recognition (CVPR)}, July 2017.

\end{thebibliography}
\bibliographystyle{plain}

\end{document}